\pdfoutput=1
\documentclass[11pt]{article}

\usepackage{emnlp2021}

\usepackage{times}
\usepackage{latexsym}
\usepackage{soul}
\usepackage{url}
\usepackage{CJK}
\usepackage{graphicx}
\usepackage{amsmath}
\usepackage{amsthm}
\usepackage{amssymb,bm}
\usepackage{amsfonts}
\usepackage{booktabs}
\usepackage{color}
\usepackage{multirow}
\usepackage{subfigure}
\usepackage{enumitem}
\usepackage{makecell}
\usepackage{amssymb}% http://ctan.org/pkg/amssymb
\usepackage{pifont}% http://ctan.org/pkg/pifont

\newcommand{\cmark}{\ding{51}}%
\newcommand{\xmark}{\ding{55}}%

\usepackage{algorithm}
\usepackage{algorithmicx}
\usepackage{algpseudocode}

\renewcommand{\algorithmiccomment}[1]{\bgroup\hfill//~#1\egroup}

\algnewcommand{\LineComment}[1]{\State \(\triangleright\) #1}

\usepackage[T1]{fontenc}
\usepackage[utf8]{inputenc}

\usepackage{microtype}

\title{A Graph-Based Neural Model for End-to-End Frame Semantic Parsing}

\author{
Zhichao Lin$^1$, Yueheng Sun$^2$, Meishan Zhang
$^{1}\thanks{~~Corresponding author.}$\\
$^1$School of New Media and Communication, Tianjin University, China\\
$^2$College of Intelligence and Computing, Tianjin University, China\\
\texttt{\{chaosmyth,yhs,zhangmeishan\}@tju.edu.cn}
}

\begin{document}
\begin{CJK}{UTF8}{gbsn}
\maketitle
\begin{abstract}
Frame semantic parsing is a semantic analysis task based on FrameNet which has received great attention recently. The task usually involves three subtasks sequentially: (1) target identification, (2) frame classification and (3) semantic role labeling. The three subtasks are closely related while previous studies model them individually, which ignores their intern connections and meanwhile induces error propagation problem. In this work, we propose an end-to-end neural model to tackle the task jointly. Concretely, we exploit a graph-based method, regarding frame semantic parsing as a graph construction problem. All predicates and roles are treated as graph nodes, and their relations are taken as graph edges. Experiment results on two benchmark datasets of frame semantic parsing show that our method is highly competitive, resulting in better performance than pipeline models.
\end{abstract}

\section{Introduction}
\begin{figure} \centering
\includegraphics[scale=0.7]{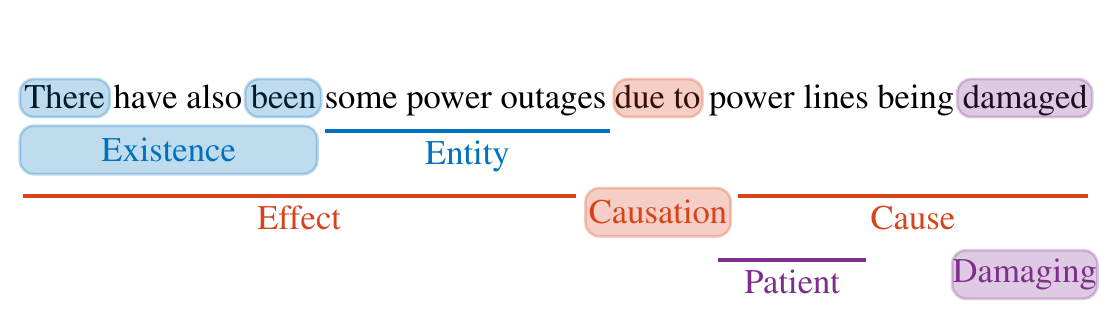}
\caption{An example involving frame semantic structures, taken from the FrameNet \cite{baker-etal-1998-berkeley}. Frame-evoking predicates are highlighted in the sentence, and corresponding frames are shown in colored blocks below. The frame-specific roles are underlined with their frames in the same row.}
\label{figure:example}
\end{figure}

Frame semantic parsing~\cite{Gildea:2002:ALS:643092.643093} aims to analyze all sentential predicates as well as their FrameNet roles as a whole, which has received great interest recently. This task can be helpful for a number of tasks, including information extraction~\cite{surdeanu-etal-2003-using}, question answering~\cite{shen-lapata-2007-using}, machine translation~\cite{liu-gildea-2010-semantic} and others~\cite{coyne-etal-2012-annotation, chen2013, agarwal-etal-2014-frame}. Figure~\ref{figure:example} shows an example, where all predicates as well as their semantic frame and roles in the sentence are depicted.

Previous studies~\cite{das-etal-2014-frame, swayamdipta2017frame, bastianelli2020encoding} usually divide the task into three subtasks, including target identification, frame classification and semantic role labeling (SRL), respectively. By performing the three subtasks sequentially, the whole frame semantic parsing can be accomplished. The majority of works focus on either one or two of the three subtasks, treating them separately~\cite{yang-mitchell-2017-joint, botschen-etal-2018-multimodal, swayamdipta-etal-2018-syntactic, peng-etal-2018-learning}.

The above formalization has two weaknesses. First, the individual modeling of the three subtasks is inefficient to utilize the relationship among them. Apparently, the earlier subtasks can not exploit the information from their future subtasks. Second, the pipeline strategy can suffer from the error propagation problem, where the errors occurring in the previous subtasks can influence the later subtasks as well. To address the two weaknesses, end-to-end modeling is one promising alternative, which has been widely adopted in natural language processing~(NLP)~\cite{cai-etal-2018-full, he-etal-2018-jointly, sun-etal-2019-joint, fu-etal-2019-graphrel, fei-etal-2020-high}. 

In this work, we propose a novel graph-based model to tackle frame semantic parsing in an end-to-end way, using a single model to perform the three subtasks jointly. We organize all predicates and their FrameNet semantic by a graph, and then design an end-to-end neural model to construct the graph incrementally. An encoder-decoder model is presented to achieve the graph building goal, where the encoder is equipped with contextualized BERT representation~\cite{devlin-etal-2019-bert}, and the decoder includes node generation and edge building sequentially. Our final model is elegant and easy to understand as a whole.

We conduct experiments on two benchmark datasets to evaluate the effectiveness of our proposed model. First, we study our graph-based framework in two settings, the end-to-end scenario and the pipeline manner, where the node building and edge building are trained separately. Results show that end-to-end modeling is much better. Besides, we also compare our model with several other pipelines, where the similar findings can be observed. Second, we compare our graph-based framework with previous methods by the three subtasks individually, finding that the graph-based architecture is highly competitive. We can obtain the best performance in the literature, leading to a new state-of-the-art result. Further, we conduct extensive analyses to understand our method in depth.

In summary, we make the following two major contributions in this work:
\setlist{nolistsep}
\begin{itemize}
  \item[(1)] We propose a novel graph-based model for frame semantic parsing which can achieve competitive results for the end-to-end task as well as the individual subtasks.
  \item[(2)] To the best of our knowledge, we present the first work of end-to-end frame semantic parsing to solve all included subtasks together in a single model.
\end{itemize}
We will release our codes as well as experimental setting public available on \url{https://github.com/Ch4osMy7h/FramenetParser} to help result reproduction and facilitate future researches.

\section{Related Work}
\paragraph{Frame-Semantic Parsing}
Frame-semantic parsing has been received great interest since being released as an evaluation task of SemEval 2007~\cite{baker-etal-2007-semeval}. The task attempts to predict semantic frame structures defined in FrameNet~\cite{baker-etal-1998-berkeley} which are composed of frame-evoking predicates, their corresponding frames and semantic roles. Most of the previous works~\cite{das-etal-2014-frame, swayamdipta2017frame, bastianelli2020encoding} focus on a pipeline framework to solve the task, training target identification, frame classification and semantic role labeling models separately. In this work, to the best of our knowledge, we present the first end-to-end model to handle the task jointly.

Among the three subtasks of frame semantic parsing, semantic role labeling has been researched most extensively \cite{kshirsagar-etal-2015-frame, yang-mitchell-2017-joint, peng-etal-2018-learning, swayamdipta-etal-2018-syntactic, marcheggiani-titov-2020-graph}. It is also highly related to the Propbank-style semantic role labeling \cite{palmer-etal-2005-proposition} as while with only differences in the frame definition. Thus the models between the two types of semantic role labeling can be mutually borrowed. There are several end-to-end Propbank-style semantic role labeling models as well \cite{cai-etal-2018-full, he-etal-2018-jointly, end2endsrl_aaai, fu-etal-2019-graphrel}. 
However, these models are difficult to be applied directly for frame semantic parsing due to the additional frame classification as well as the discontinuous predicates. In this work, we present a totally-different graph construction style model to solve end-to-end frame semantic parsing elegantly.

\paragraph{Graph-Based Methods}
Recently, graph-based methods have been widely used in a range of other tasks, such as dependency parsing~\cite{dozat2016deep, kiperwasser-goldberg-2016-simple, ji-etal-2019-graph}, AMR parsing~\cite{flanigan-etal-2014-discriminative, lyu-titov-2018-amr, zhang-etal-2019-amr, zhang-etal-2019-broad} and relation extraction~\cite{sun-etal-2019-joint, fu-etal-2019-graphrel, dixit-al-onaizan-2019-span}. In this work, we aims for frame semantic parsing, organizing the three included subtasks by a well designed graph, converting it into graph-based parsing task naturally.

\section{Method}
% In this section, we first formally define the task of end-to-end frame-semantic parsing in Section 3.1. Next, we bring in word representation representation that our method based on in section 3.2. Then we will introduce our graph-based method in Section 3.3. Finally, we extend our method to joint setting in Section 3.4 and describe the  decoding procedure in Section 3.5.

\begin{figure*} \centering
  \includegraphics[scale=0.77]{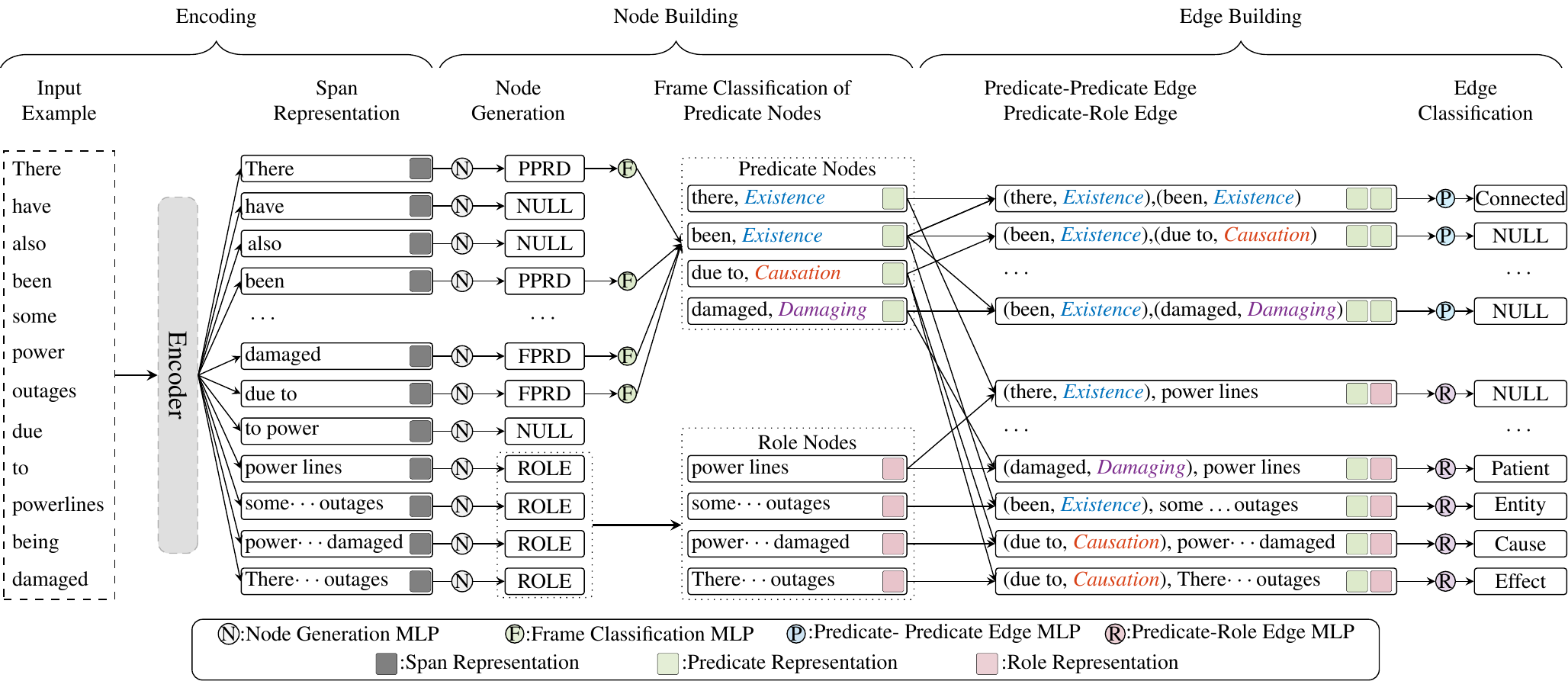}
  \caption{The overall architecture of our graph-based end-to-end model.}
  \label{figure:model}
  \end{figure*}

\subsection{Task Formulation}
The goal of frame-semantic parsing is to extract semantic predicate-argument structures from texts, where each predicate-argument structure includes a predicate by a span of words, a well-defined semantic frame to express the key roles of the predicate, and the values of these roles by word spans. Formally, given by a sentence $X$ with $n$ words $w_1, w_2, \dots, w_n$, frame-semantic parsing aims to output a set of tuples $\mathcal{Y} = \{(y_1, y_2, \dots, y_k)\}_{k=1}^{K}$, where each $y_i$ consists of the following elements:
\setlength\itemsep{0em}
\begin{itemize}
  \item $p_{i} = (p_{i,1}, \dots, p_{i,d_i})$, where $p_{i,*}$ are word spans in X and $d_i$ indicates the number of pieces of the predicate since it might be discontinuous.
  \item $f_i \in \mathcal{F}$, where $\mathcal{F}$ is the frame set which is well defined in FrameNet.\footnote{The Berkeley FrameNet~\cite{baker-etal-1998-berkeley} project provides a lexicon of semantic frames. Specifically, the FrameNet 1.5 is with 1020 lexicalized frames, while in 1.7, the number is 1221.}
%   By using this type, we can solve the overlapped problem between common targets and discontinuous targets.
  \item $r_i = ([r_{i,1},v_{i,1}], \dots, [r_{i, m_k},v_{k,m_k}])$, where $r_{i,*}$ are frame roles derived from $f_i$ and $v_{i ,*}$ are also word spans in $X$.
\end{itemize}
The full frame semantic parsing is usually divided into the following three subtasks:
\begin{itemize}
\setlength\itemsep{0em}
    \item \textbf{Target Identification} (also known as predicate identification), which is to identify all valid frame-evoking predicates from $X$, outputting $P = \{(p_1, ..., p_k)\}$.
    \item \textbf{Frame Classification}, which is to predicate the concrete evoking frame  $f_i$ of a certain predicate $p_i \in P$.
    \item \textbf{Semantic Role Labeling}, which is to assign concrete values for roles  $r_i$ by given a predicate frame pair $(p_i, f_i)$.
\end{itemize} 
Previously, the majority of work of frame semantic parsing performs the three subtasks individually, ignoring their highly-related connections and also being vulnerable to the error propagation problem. Thus, we present an end-to-end graph-based model to accomplish the three subtasks by a single model.

%In this work, we focus on end-to-end frame-semantic parsing, assume no gold annotations are provided. Our work attempts to extract all possible frame semantic structures directly from one sentence, instead of applying the flat paradigm(one instance with one \textit{target}).

%\subsection{Word Representation}
%Due to the powerful capabilities of BERT\cite{devlin-etal-2019-bert}, we adopt it as the input of our model. Given a sentence $X=\{w_{1},w_{2},...,w_{n}\}$, BERT will convert each word $w_{i}$ into word pieces, and feed them into deep transformer encoders to get the piece-level representation. To obtain word-level representation, we average all pieces vector representations of word $w_{i}$ as its final representation $\mathbf{e}_{i}$. Following \cite{swayamdipta-etal-2018-syntactic}, we exploit BiHSLTM\cite{NIPS2015_215a71a1} to convert the word representation $\mathbf{e}_{i}$ into contextual representation:
%\begin{equation} 
%    \label{lstm}
%    \mathbf{h}_1, \cdots, \mathbf{h}_n = \text{BiHLSTM}(\mathbf{e}_1, \cdots, \mathbf{e}_n)
%\end{equation}

\subsection{The Graph-Based Methodology}
\label{sec:method}
We formalize the frame-semantic parsing task as a graph constructing problem, and further present an encoder-decoder model to perform the task in an end-to-end way. The encoder aims for representation learning of the frame semantic parsing, and the decoder constructs the semantic graph incrementally. Concretely, for the encoder, we compute the span representations since the basic processing units of our model are word spans, and for the decoder, we first generate all graph nodes, and then build edges among the graph nodes. Figure~\ref{figure:model} shows the overall architecture of our method.
\subsubsection{Encoding}

Due to the strong capability of BERT~\cite{devlin-etal-2019-bert} for represent learning, we adopt it as the backbone of our model. Given a sentence $X=\{w_{1},w_{2},...,w_{n}\}$, BERT converts each word $w_{i}$ into word pieces, and feed them into deep transformer encoders to get the piece-level representation. To obtain word-level representation, we average all piece vectors of word $w_{i}$ as its final representation $\mathbf{e}_{i}$. 

For further feature abstraction, we exploit BiHLSTM~\cite{NIPS2015_215a71a1} to compose high-level features based on word-level output $\mathbf{e}_1, \cdots, \mathbf{e}_n$, following~\citet{swayamdipta-etal-2018-syntactic}:
\begin{equation} 
    \label{lstm}
    \mathbf{h}_1, \cdots, \mathbf{h}_n = \text{BiHLSTM}(\mathbf{e}_1, \cdots, \mathbf{e}_n),
\end{equation}
where the gated highway connections are applied to BiLSTMs.

\paragraph{Span Representation}
We enumerate all possible spans $S=\{s_1,s_2,\dots,s_m\}$ in a sentence and limit the maximum span length to L. Then, each span $s_i \in S$ is represented by:
\begin{equation}
    \mathbf{g}_i = [\mathbf{h}_{\textsc{START}(i)}; \mathbf{h}_{\textsc{END}(i)}; \mathbf{h}_{\textsc{attn}}; \phi(s_i)],
\end{equation}
where $\phi(g_i) $ represents the learned embeddings of span width features, $\mathbf{h}_{\textsc{attn}}$ is computed by self-attention mechanism which weights the corresponding vector representations of the words in the span by normalized attention scores, and $\textsc{START}(i)$ and $\textsc{END}(i)$ denote start and end indices of $s_i$.

\subsubsection{Node Building}
\label{sec: nodebuild}
\paragraph{Node Generation}
We exploit a preliminary classification to achieve the goal of node generation. First, a span can be either a graph node or not. Further, a graph node can be a full or partial predicate node, and the node can also be a role node. Totally, we define four types for a given span:
\begin{itemize}
  \item \texttt{FPRD}: a full predicate span.
  \item \texttt{PPRD}: a partial predicate span.
  \item \texttt{ROLE}: a role span.
%   By using this type, we can solve the overlapped problem between common targets and discontinuous targets.
 \item \texttt{NULL}: a span that is not a graph node.
\end{itemize}
The type of a span can be the full permutation of elements in set $\{\texttt{FPRD}, \texttt{PPRD}, \texttt{ROLE}\}$ or $\texttt{NULL}$. Thus, each span can be classified into eight types (i.e., \texttt{FPRD}, \texttt{PPRD}, \texttt{ROLE}, \texttt{FPRD-PPRD},  \texttt{FPRD-ROLE}, \texttt{PPRD-ROLE}, \texttt{FPRD-PPRD-ROLE}, \texttt{NULL}). 

Given an input span $s_i$ with its vectorial representation as $\mathbf{g}_i$, we exploit one MLP layer with softmax to classify the span type:
\begin{equation} \label{eq:tscore}
 \mathbf{p}_n = \mathrm{softmax}(\mathrm{MLP}_n(\mathbf{g}_i)),
\end{equation}
where $\mathbf{p}_n$ indicates the probabilities of span types. By this classification, all non-null type spans are graph nodes, reaching the goal of node generation.
 
\paragraph{Frame Classification of Predicate Nodes}
Node generation detects all graph nodes roughly, assigning each node with a single label to indicate whether it can be served as a predicate or role. Here we go further to recognize the semantic frames for all predicate nodes, which could be regarded as an in-depth analysis for node attribution. The step is corresponding to the frame classification subtask. 

Given an input span $s_i$ of a predicate node (\texttt{FPRD} or \texttt{PPRD}),  assuming its representation being $\mathbf{g}_i$, 
we use another MLP layer together with softmax to output the probabilities of each candidate frame for the predicate node:
\begin{equation} \label{eq:fscore}
 \mathbf{p}_c = \mathrm{softmax}(\mathrm{MLP}_c(\mathbf{g}_i)),
\end{equation}
where $\mathbf{p}_c$ is the output probabilities of semantic frames. Specially, frames are constrained by the lexical units defined in FrameNet. For example, the predicate with the lexical unit "meeting" only evokes frame \textit{Social\_event} and \textit{Discussion}.

We also adopt the pseudo strategy following ~\citet{swayamdipta2017frame} to optimize the classification. First, we use spacy lemmatizer~\cite{spacy} to translate an input sentence into lemmas. Then, if a word span is a predicate node, we treat the corresponding lemma span as the pseudo lexical unit and index the corresponding semantic frame set by it. Finally, we reduce the search space by masking frames outside the set. In our experiments, we find it is practical to apply this strategy.

\subsubsection{Edge Building}
\label{sec: edgebuild}
After graph nodes are ready, we then build edges to accomplish frame semantic parsing accordingly. There are two types of edges in our model.

\paragraph{Predicate-Predicate Edge}
For extracting discontinuous mentions, we build the edges between nodes which are predicate fragments (i.e., \texttt{PPRD} nodes). In detail, we treat it as a binary classification problem considering whether two nodes alongside the edge can form parts of a predicate or not. Formally, given two \texttt{PPRD} nodes with the corresponding spans $\mathbf{s}^p_i$ and $\mathbf{s}^p_j$ and their encoding representations $\mathbf{g}^p_i$ and $\mathbf{g}^p_j$, we utilize one MLP layer to classify their edge type:
\begin{equation}
\label{eq:pred_rel}
    \mathbf{p}_{pe} = \mathrm{softmax}(\mathrm{MLP}_{pe}([\mathbf{g}^p_i, \mathbf{g}^p_j, \mathbf{g}^p_i*\mathbf{g}^p_j])),
\end{equation}
where $\mathbf{p}_{pe}$ indicates the probabilities of two types, namely \texttt{Connected} and \texttt{NULL} (i.e., cannot be connected), and the feature representation is borrowed from~\citet{zhao-etal-2020-spanmlt}.

\paragraph{Predicate-Role Edge}
For extracting frame-specific roles, we build the edges between predicates nodes (i.e., node type by \texttt{FPRD} or \texttt{PPRD}) and role nodes (i.e., node type by \texttt{ROLE}). Given a predicate node $\mathbf{s}^p_i$ and a role node $\mathbf{s}^r_j$, assuming their neural representations being $\mathbf{g}^p_i$ and $\mathbf{g}^r_j$, respectively, we utilize another MLP layer to determine their edge type by multi-class classification:
\begin{equation}
\label{eq:role_rel}
    \mathbf{p}_{re} = \mathrm{softmax}(\mathrm{MLP}_{re}([\mathbf{g}^p_i, \mathbf{g}^r_j,\mathbf{g}^p_i*\mathbf{g}^r_j])),
\end{equation}
where $\mathbf{p}_{re}$ indicates the probabilities of predicate-role edge types (i.e., frame roles as well as a \texttt{NULL} label indicating no relation). 

\subsection{Joint Training}
\label{sec:joint}
To train the joint model, we employ the negative log-likelihood loss function for both node building and edge building step:
\begin{equation} 
\label{eq:loss_part}
\begin{split}
 \mathcal{L}_{n} &= - \sum \log \mathbf{p}_n(y_n) - \sum \log \mathbf{p}_c(y_c)\\
 \mathcal{L}_{e} &= - \sum \log \mathbf{p}_{pe}(y_{pe}) - \sum \log \mathbf{p}_{re}(y_{re}), \\
\end{split}
\end{equation}
where $y_n$ and  $y_c$ are the gold labels for the text spans and predicate nodes,
$y_{pe}$ and $y_{re}$ indicate the gold edge labels for the predicate-predicate and predicate-role node pairs.
Further, losses from two steps are summed together, leading to the final training objective of our model:
\begin{equation} 
\label{eq:loss_part}
 \mathcal{L} = \mathcal{L}_{n} + \mathcal{L}_{e}
\end{equation}

\subsection{Decoding}
\label{sec:decoding}
The decoding aims to derive frame semantic parsing results by the graph-based model. Here we describe the concrete process by the three subtasks.

\paragraph{Target Identification}
The target identification involves both node building and edge building steps. First, all predicate nodes with type \texttt{FPRD} are predicates. Second, there is a small percentage of predicates composed of multiple nodes with type \texttt{PPRD}. If two or more such nodes are connected with predicate-predicate edges, we regard these nodes as one single valid predicate .

\paragraph{Frame Classification}
The frame classification decoding is performed straightforwardly for single-node predicates. For multi-node predicates, there may exist conflicts from the frame classification of different nodes. Concretely, given a multi-node predicate composed of two or more nodes, the max-scored frame evoked by them might be different. Thus, to address this issue, we use the maximum operation achieved by first summing up the softmax distributions over all covered nodes and then fetching the max-scored frame.

\paragraph{Semantic Role Labeling}
The condition of semantic role labeling is similar to frame classification. For the single-node predicates, the semantic role labeling output is determinative. For the multi-node predicates, we assign role values for the candidate roles inside its predicted frame only, and further select the concrete role node, which is the highest-probability to the covered predicate nodes.

\section{Experiments}
\subsection{Setting}
\paragraph{Dataset}
We adopt the FrameNet versions 1.5 and 1.7\footnote{\url{https://framenet.icsi.berkeley.edu/fndrupal/}} (denoted by FN1.5 and FN1.7 for short, respectively) as the benchmark datasets to evaluate our models. FN1.5 is the widely used dataset in previous work and FN1.7 is the latest version used recently which involves more semantics. We follow the previous studies~\cite{das-etal-2014-frame, swayamdipta2017frame} to divide the two datasets into the training, validation and test sets, respectively. Table~\ref{table:dataset} shows the overall data statistics.

\begin{table}[!t]
 \resizebox{0.45\textwidth}{!}{
\begin{tabular}{clccc}
\hline
\textbf{Dataset}      & \textbf{Type} & \textbf{Train} & \textbf{Dev} & \textbf{Test} \\ \hline
\multirow{3}{*}{FN1.5} & \# Sentence      &  2,713 &  326 &  982    \\ 
                      & \# Predicate    &   16,618    &  2,282 &  4,427   \\ 
                      & \# Role         &  29,449     &  4,039 &  7,146  \\ 
                      \hline
\multirow{3}{*}{FN1.7} & \# Sentence      &  3,413 & 326 & 1,354      \\ 
                      & \# Predicate    &  19,384     &  2,270   &  6,714    \\
                      & \# Role         & 34,385      &  4,024   &  11,303    \\ \hline
\end{tabular}}
\caption{Statistics of the datasets.}
\label{table:dataset}
\end{table}

\begin{table*}[!t]
\centering
\resizebox{1.0\textwidth}{!}{
  \begin{tabular}{c|l|ccc|ccc|ccc}
    \toprule
    \multirow{2}{*}{Data} & \multirow{2}{*}{Model} & \multicolumn{3}{c}{Target}  & \multicolumn{3}{c}{Frame} & \multicolumn{3}{c}{Role} \\  
    & & P & R & F1 & P & R & F1 & P & R & F1 \\
    \hline 
    \multirow{6}{*}{FN1.5} & Predicate+Frame+Role& 73.17 & 75.47 & 74.30 & 66.08 & 68.15 & 67.06 &45.91& 46.70 & 46.30  \\
    & Predicate$\circ$Frame+Role& 74.32 & \bf 76.16 & 75.23 & 67.91 & 68.78 & 68.34 & \bf 46.85 & 47.89 & 47.36 \\
    & Predicate+Frame$\circ$Role&  73.17&   75.47&  74.30 &  67.73&  68.56 &  68.14 & 46.51 & 49.01 & 47.72  \\
    & Predicate$\circ$Frame+Semi-CRF& 74.32 & \bf 76.16 & 75.23 & 67.91 & 68.78 & 68.34 & 46.33 & \bf 50.87 & \bf 48.50 \\
    & Node+Edge& \bf 74.99 & 75.85 & \bf 75.42 & \bf 68.01 & \bf 68.79 & \bf 68.40 & 46.64 & 49.35 & 47.96 \\
    \cline{2-11}
    &\bf  Ours-Joint & \bf 75.81 & \bf 76.17 & \bf 75.99 & \bf  68.72 & \bf 69.05 & \bf 68.89 & \bf 47.79 & \bf 50.60 & \bf 49.16 \\
    \hline \hline
    \multirow{6}{*}{FN1.7} & Predicate+Frame+Role& \bf 78.62 & 69.92 & 74.02 & \bf 71.05 & 63.06 & 66.82 & 49.32& 45.60 &  47.38\\
    & Predicate$\circ$Frame+Role& 76.79 & \bf 72.92 & \bf 74.81 & 69.29 &  66.16 & \bf 67.69 & 49.21 & 46.03 & 47.57 \\
    & Predicate+Frame$\circ$Role&  \bf 78.62&   69.92&  74.02 &  69.79& 65.06 &  67.34 & 49.46 & 46.01 & 47.67\\
    & Predicate$\circ$Frame+Semi-CRF& 76.79 & \bf 72.92 & \bf 74.81 & 69.29 &  66.16 & \bf 67.69 & 49.03 & \bf 47.49 & \bf 48.24 \\
    & Node+Edge& 76.81 & 72.54 & 74.62 &  68.99 &  \bf 66.33 & 67.63 & \bf 49.55 & 46.28 & 47.86  \\
    % % \hline
    \cline{2-11}
    & \bf Ours-Joint & 76.16 & \bf 74.98 & \bf 75.56 & 69.39 & \bf 68.30 & \bf 68.84 & 49.09 & \bf 48.81 & \bf 48.95 \\
    \bottomrule
  \end{tabular}}
  \caption{Main results of frame-semantic parsing on FN1.5 and FN1.7, where the pipeline and end-to-end methods are compared thoroughly.} \label{table:result}
\end{table*}

\paragraph{Evaluation}
\label{sec:eval}
We measure the performance of frame semantic parsing by its three subtasks, respectively. For target identification, we treat a predicate as correct only when all its included word spans exactly match with the gold-standard spans of the predicate. For frame classification, we use the joint performance for evaluation, regarding a classification as correct only when the predicate, as well as the frame, are both correct. For semantic role labeling, we also use the joint performance regarding the role as correct when the predicate, role span (exact match), and role type are all correct, which is treated as our major metric.

\paragraph{Derived Models}
\label{sec:baseline}
Following previous studies and our graph-based method, we can derive a range of basic models for comparisons:
\setlist{nolistsep}  
\begin{itemize}
\setlength\itemsep{0em}
  \item \textbf{Node}, the node building submodel which is the first step of our decoder module mentioned in section~\ref{sec: nodebuild}.
  \item \textbf{Edge}: the edge building submodel which is the second step of our decoder module mentioned in section~\ref{sec: edgebuild}.
  \item \textbf{Predicate}, a graph-based predicate identification model, which is implemented by keeping only the predicate node generation and predicate-predicate edge building in our final graph-based model. 
  %The model could be highly related to the target identification component as mentioned in Section~\ref{sec:decoding}. 
  \item \textbf{Frame}, our final graph-based model with only the frame classification submodel, assuming predicate nodes and their edges are given.
  \item \textbf{Role}, our graph-based model with only role node generation and predicate-role edge building, assuming predicate nodes and their frames are given.
  \item \textbf{Predicate$\circ$Frame}, a joint model of Predicate and Frame, which is implemented by excluding the role node generation and predicate-role edge building in our final model.
  \item \textbf{Frame$\circ$Role}, a joint model of Frame and Role, which is implemented by excluding the predicate node generation and predicate-predicate edge building in our final model.
  \item \textbf{Semi-CRF}, a span-level semi-Markov CRF~\cite{NIPS2004_eb06b9db} model for semantic role labeling which is borrowed from \citet{swayamdipta-etal-2018-syntactic}, where the only difference is that we use BERT as the representation layer for fair comparisons.\footnote{According to the preliminary experiments, we find that the fine-tuned method of BERT usage would hurt the Semi-CRF model performance. Therefore, we freeze the BERT parameters for Semi-CRF here.}
\end{itemize}
Note that the above derived models are trained individually. Based on these  models, we can build five pipeline systems: (1) \textbf{Predicate + Frame + Role}, (2) \textbf{Predicate$\circ$Frame + Role}, (3) \textbf{Predicate + Frame$\circ$Role}, (4) \textbf{Predicate$\circ$Frame + Semi-CRF}, and (5) \textbf{Node + Edge},  which are exploited for comparisons with our graph-based end-to-end model.

\paragraph{Hyperparameters}
All our codes are based on Allennlp Library \cite{Gardner2017AllenNLP} and trained on a single RTX-2080ti GPU. We choose the BERT-base-cased\footnote{https://github.com/google-research/bert}, which consists of 12-layer transformers with the hidden size 768 for all layers. We set all the hidden sizes of BiHLSTM to 200, and the number of layer to 6. The MLP layers are of dimension size by 150 and depth by 1, with ReLU function. We apply dropouts of 0.4 to BiHLSTM and 0.2 to MLP layers. Following \citet{swayamdipta-etal-2018-syntactic}, we also limit the maximum length of spans to 15 for efficiency, resulting in oracle recall of 95\% on the development set.

For training, we exploit online batch learning with a batch size of 8 to update the model parameters, and use the BertAdamW algorithm with the learning rate $1 \times 10^{-5}$ to finetune BERT and $1 \times 10^{-3}$ to fine-tune other parts of our model. The gradient clipping mechanism by a maximum value of 5.0 is exploited to avoid gradient explosion.
The training process are stopped early if the performance does not increase by 20 epochs.

\subsection{Main Results}
\label{sec:result}
Table \ref{table:result} shows the main results on the test sets of FN1.5 and FN1.7 datasets respectively, where our end-to-end model is compared with the four strong pipeline methods mentioned in Section~\ref{sec:baseline}. We can see that the end-to-end joint model can lead to significantly better performance (p-value below $10^{-5}$ by pair-wise t-test) as a whole on both datasets. Concretely, we can obtain average improvements of  $\frac{0.57 + 0.75}{2} = 0.66 $ on target identification, $\frac{0.49 + 1.15}{2} = 0.82$ on frame classification, and  $\frac{0.66 + 0.71}{2} = 0.69$ on semantic role labeling on the two datasets compared with the best results of the pipeline systems, respectively.  

Besides the overall advantage of the end-to-end joint model over the pipelines,
we can also find that the joint of two subtasks can also outperform their counterpart baselines.
Concretely, as shown in Table \ref{table:result}, Predicate$\circ$Frame is better than Predicate + Frame,
and Frame$\circ$Role is better than Frame + Role.
The results further indicate the effectiveness of joint learning.
Further, by comparisons between our graph-based Role model and the Semi-CRF one,
we can see that the Semi-CRF is better.
The reason could be that the Semi-CRF model can exploit higher-order features among different frame roles which are ignored by our simple edge building module.  
As our edge building considers all predicates and all roles together, 
the incorporation of such features is still with great inconveniences.

\begin{table}[!t]
  \begin{center}
  \resizebox{0.45\textwidth}{!}{
  \begin{tabular}{l c c}
      \toprule
      Model & FN1.5 & FN1.7 \\
      \midrule
      \citet{das-etal-2014-frame} & 45.40 & -  \\
      \citet{swayamdipta2017frame} & 73.23 & \bf 73.25 \\
      \citet{bastianelli2020encoding} (wo syntax) & 74.96 & - \\
      \citet{bastianelli2020encoding} (w syntax) & \bf 76.80 & - \\
      \midrule
      Predicate & 76.09 & 75.34 \\
      Predicate$\circ$Frame & 76.47  & 75.88  \\
      Ours-Joint & \bf 76.90 & \bf 76.27\\
      \bottomrule
  \end{tabular}}
  \caption{Target Identification Results.}
  \label{table:target}
\end{center}
\end{table}

\begin{table}[!t]
  \begin{center}
  \resizebox{0.45\textwidth}{!}{
  \begin{tabular}{l c c}
      \toprule
      Model & FN1.5 & FN1.7 \\
      \midrule
      \citet{das-etal-2014-frame} & 83.60 & -  \\
      \citet{hermann-etal-2014-semantic} & 88.41 & - \\
      \citet{hartmann-etal-2017-domain} & 87.63 & - \\
      \citet{yang-mitchell-2017-joint} & 88.20 & - \\
      \citet{swayamdipta2017frame} & 86.40 & 86.55 \\
      \citet{botschen-etal-2018-multimodal} &  88.82 & - \\
      \citet{peng-etal-2018-learning} & \bf 90.00 & \bf 89.10 \\
      \citet{bastianelli2020encoding} (wo syntax) & 89.90 & - \\
      \citet{bastianelli2020encoding} (w syntax) & 89.83 & - \\
      \midrule
      Frame & 90.16 & 90.34 \\
      Ours-Joint & \bf 90.62 & \bf 90.64 \\
      \bottomrule
  \end{tabular}}
  \caption{The accuracy of the Frame Identification task based on the gold targets.}
  \label{table:frame}
\end{center}
\end{table}

\begin{table}[!t]
  \begin{center}
  \resizebox{0.45\textwidth}{!}{
  \begin{tabular}{l c c}
      \toprule
      Model & FN1.5 & FN1.7 \\
      \midrule
      \citet{das-etal-2014-frame} & 59.10 & -  \\
      \citet{kshirsagar-etal-2015-frame} & 63.10 & - \\
      \citet{yang-mitchell-2017-joint} & 65.50 & - \\
      \citet{swayamdipta2017frame} & 59.48 & \bf 61.36 \\
      \citet{swayamdipta-etal-2018-syntactic} & 69.10 & - \\
      \citet{marcheggiani-titov-2020-graph} & 69.30 & - \\
      \citet{bastianelli2020encoding} (wo syntax) &  72.85 & - \\
      \citet{bastianelli2020encoding} (w syntax) & \bf \bf 75.56 & - \\
      \midrule
      Semi-CRF & \bf 73.56 & \bf 72.22 \\
      Ours-Joint & 73.28 & 72.06 \\
      \bottomrule
  \end{tabular}}
  \caption{Pipeline Semantic Role Labeling results using gold targets and frames.}
  \label{table:frame-element}
\end{center}
\end{table}

\subsection{Individual Subtask Evaluation}
Previous studies commonly focus on only individual subtasks of frame semantic parsing. In order to compare with these studies, we simulate the scenarios by imposing constraints with gold-standard inputs in our joint models. In this way, we show the capability of our models on individual tasks.\footnote{The results are obtained by the scripts from \citet{swayamdipta2017frame}.} In particular, \citet{bastianelli2020encoding} report the best performance of the previous studies in the literature, which is based on BERT representations. They adopt the constituency syntax which can boost the individual model performances significantly. 
Since our final model uses no other knowledge except BERT, we report their model performances by with syntax (denoted as w syntax) and without syntax (denoted as wo syntax) for careful comparisons. 

\paragraph{Target Identification}
We show the performance of previous studies on target identification in Table~\ref{table:target}, and also report the results of three-related models derived from this work. 
First, by comparing our three models (i.e. predicate only, predicate with frame, and the full graph parsing), the results show that both frame classification and semantic role labeling can help target identification.
Second, we can see that our final model can achieve the best performance among the previous work. 

\paragraph{Frame Classification}
Noted that we have 
Table~\ref{table:frame} shows the result of individual frame classification tasks, where all systems assume gold-standard predicates as inputs. Similar to target identification, we can achieve better performance than all previous studies. \citet{peng-etal-2018-learning} did not use BERT, but they use extra datasets from FrameNet (exemplar sentences) and semantic dependency parsing, which can also benefit our task greatly. As for the comparison between our implemented two models, Frame alone and our final joint model, the results show that semantic role labeling can benefit the frame classification, which is reasonable.

\paragraph{Semantic Role Labeling}
Table~\ref{table:frame-element} shows the results of various models on the semantic role labeling task. By constraining gold-standard predicates and frames to the outputs, our model degenerates to a normal semantic role labeling model. We also give the result by using Semi-CRF. As shown, our final semantic role labeling model is highly competitive in comparison with previous studies, except the model of \citet{bastianelli2020encoding} with syntax.  The exception is expected, since syntax has been demonstrated highly effective before for SRL \cite{swayamdipta-etal-2018-syntactic, peng-etal-2018-learning, bastianelli2020encoding}.  
In addition, the Semi-CRF model is better than our method, which is consistent with the results in Table \ref{table:result}.

\begin{figure}
\centering
\subfigure[FN1.5]{\label{figure:ks_1}
\includegraphics[scale=0.49]{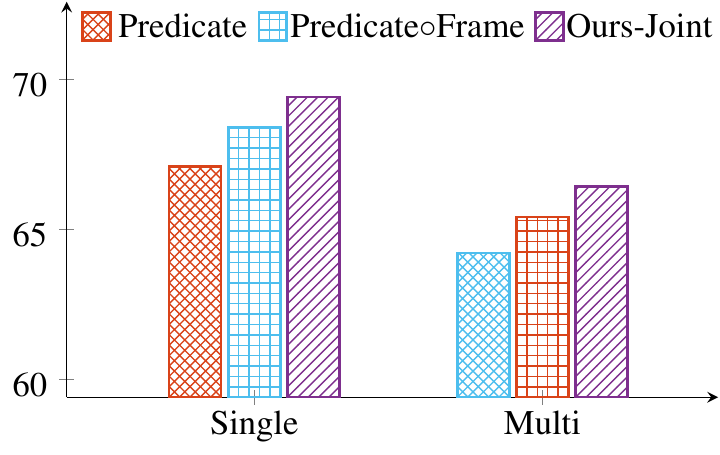}
}
\subfigure[FN1.7]{\label{figure:ks_2}
\includegraphics[scale=0.49]{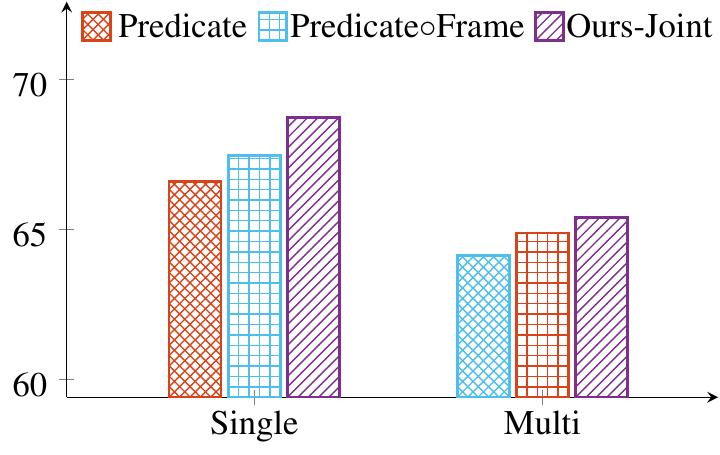}
}
\caption{F1 scores of frame metric in different predicates. \textbf{Single}: single-word predicate. \textbf{Multi}: multi-word predicate.}
\label{figure:pred-f1}
\end{figure}

 \subsection{Discussion}
In this subsection, we conduct detailed experimental analyses for better understanding our graph-based methods. Note that if not specified, the analyses are based on the FN1.7 dataset, which has a larger scale of annotations for exploring.

\begin{table}[t]
  \centering
  \resizebox{0.47\textwidth}{!}{
  \begin{tabular}{l|ccc}
  \toprule
  Model & Node & Frame & Edge  \\
  \midrule
  Predicate+Frame+Role & 64.17 &  68.39 & 50.14\\
  Predicate$\circ$Frame+Role & 64.32 &  68.89 & 50.97\\
  Predicate+Frame$\circ$Role & 64.24 &  68.97 & 51.09\\
  Node+Edge & \bf 64.56 & \bf 69.30 & \bf 51.43\\ \hline
  Ours-Joint & \bf 65.34 & \bf 69.80 & \bf 52.13\\
  \bottomrule
  \end{tabular}}
  \caption{F1 score results of the modules.}
  \label{table:node-result}
  \end{table}

\begin{table}[t]
  \centering
  \resizebox{0.47\textwidth}{!}{
  \begin{tabular}{l|ccc}
  \toprule
  Model & Target & Frame & Role  \\
  \midrule
  Predicate+Frame+Role & 74.86 & 67.28 & 47.34\\
  Predicate$\circ$Frame+Role & \bf 75.11 & \bf 67.78 & 47.81\\
  Predicate+Frame$\circ$Role & 74.86 & 67.57 & 47.93 \\
  Predicate$\circ$Frame+Semi-CRF & \bf 75.11 & \bf 67.78 & \bf 48.26 \\
  \hline
  Ours-Joint & \bf 75.72 & \bf 68.86 & \bf 48.89\\
  \bottomrule
  \end{tabular}}
  \caption{F1 score results of our proposed method ignoring discontinuous predicates.}
  \label{table:wo-disc}
\end{table}

\paragraph{Effectiveness on recognizing different types of predicates}
For frame-semantic parsing, extracting correct frame-evoking predicates is the first step that influences the later subtasks directly. Here we performed fine-grained analysis for the predicate identification, splitting the predicates into three categories, i.e., single-word predicates (Single), multi-word predicates (Multi), respectively. As shown in Figure \ref{figure:pred-f1}, our joint model can achieve consistent improvements over the pipeline models for all kinds of predicates, 
indicating that the information from the frame and frame-specific roles are both beneficial for target identification. 
In addition, the multi-word predicates are more difficult than the single-word predicates, leading to significant decreases as a whole.

\begin{figure}[t]\centering
	\includegraphics[scale=1.0]{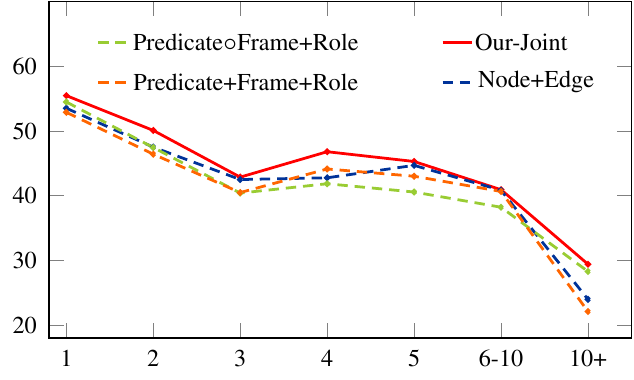}
	\caption{F1 scores of roles in terms of the length.}
	\label{figure:role-f1}
\end{figure}

\setlength{\tabcolsep}{6pt}
\begin{table}[t]
\small
\begin{center}
    \resizebox{0.4\textwidth}{!}{
    \begin{tabular}{lccc}
    \toprule
        Model
        & FN1.5
        & FN1.7\\
    \midrule
        Semi-CRF & 1.94 sent/s & 1.72 sent/s  \\
        Ours-Joint & \bf 15.72 sent/s & \bf 16.51 sent/s \\ 
    \bottomrule
    \end{tabular}}
{
    \caption{Comparison on decoding speed (sentences per second) for the semantic role labeling subtask.}
    \label{table:computational_eff}
} 
\end{center}
\end{table}

\renewcommand\arraystretch{1.0}
\setlength{\tabcolsep}{3pt}
\definecolor{a1}{HTML}{0F0F00}
\definecolor{a2}{HTML}{E56709}
\begin{table} [h]\centering \small
\begin{tabular}{c|p{6.2cm}} % {p=6.8cm}
\toprule
\bf \makecell{LU \\ Lexicon} & \multicolumn{1}{c}{\bf Text and Frames} \\  % \cline{2-10}
\hline
\multirow{3}{*}{\cmark}   &
Up Tai Hang \textcolor{a1}{$[${\bf Road}$]_{Roadways}$} behind Causeway Bay is Aw Boon Haw (Tiger Balm) \textcolor{a1}{$[${\bf Gardens}$]_{Locale\_by\_use}$}. \\
\hline
\multirow{3}{*}{\xmark}    &
Up Tai Hang \textcolor{a1}{$[${\bf Road}$]_{Roadways}$} \textcolor{blue}{$[${\bf behind}$]_{Locative\_relation}$} Causeway Bay is Aw Boon Haw (Tiger Balm) \textcolor{a1}{$[${\bf Gardens}$]_{Locale\_by\_use}$}.  \\
\hline
\hline
\multirow{3}{*}{ \makecell{Ground \\ Truth}} &
Up Tai Hang \textcolor{a1}{$[${\bf Road}$]_{Roadways}$} behind Causeway \textcolor{a1}{$[${\bf Bay}$]_{Natural\_features}$} is Aw Boon Haw (Tiger Balm) \textcolor{a1}{$[${\bf Gardens}$]_{Locale\_by\_use}$}. \\
\bottomrule
\end{tabular}
\caption{An example for frame suggestion out-the-scope-of the predefined LU lexicon, where the blue indicates the suggested frame outside the dictionary, \cmark and \xmark~represent whether inference with the dictionary, respectively.}

\label{tab:case:study}
\end{table}

\paragraph{Performance by the role length}
Frame-special roles are the core structures that frame-semantic parsing intends to obtain. It is obvious that roles of different lengths would affect the performance, and longer roles would be much more difficult. Here we bucket the roles into seven categories and report the F1-score of our proposed methods on them. Figure~\ref{figure:role-f1} shows the results.  We can find that the overall curve declines as the length increases, which is consistent with our intuition, and our graph-based end-to-end model is better than the pipeline methods of all lengths.

\paragraph{Performances of node, frame and edge}
Our graph-based model builds nodes, determines node attributes (frame), and builds edge sequentially,  which is different from the standard pipelines based on target identification, frame classification and semantic role labeling.
Thus, it is interesting to see the performance based on node building, frame classification\footnote{The frame classification is slightly different from that in the pipeline systems, which also includes the frame classification of the \texttt{PPRD} nodes. } and edge building, respectively. 
Table~\ref{table:node-result} shows the results,
where the joint model as well as four pipeline models are included.
As shown, we can see that the full joint model is better than the partial joint models,
and the full pipeline model gives the worst results.  

\paragraph{Ignoring discontinuous predicates}
Although in both FN1.5 and FN1.7 datasets, discontinuous predicates are significantly smaller in amount than others, we keep it in this work for a more comprehensive study to demonstrate that our model can process them as well. Here we also add the results which ignore the discontinuous predicates (i.e., removing the predicate- predicate edges) to facilitate future studies. As shown in Table~\ref{table:wo-disc}, our joint model performs better than the pipeline methods, which is consistent with the main results.

\paragraph{Comparison on decoding speed}

% Experimental results are all obtained by running models on a single 2080ti GPU. 
Table 8 compares the computational efficiency of the strong Semi-CRF baseline and our joint model for semantic role labeling task, which is also an essential measurement of proposed approach. Experimental results are all obtained by running models on a single 2080ti GPU. We could observe that our model can reach an almost ten times faster speed in comparison to Semi-CRF. Even though the Semi-CRF implementation\footnote{\url{https://github.com/swabhs/scaffolding}.} uses dynamic programming to optimize the time complexity, it still needs to iterate over segments of each sentence in the batch one by one, which might not take advantage of the GPU's parallel capabilities to accelerate the process. Nevertheless, our model as a whole adopts batch-based learning, which enables more efficient inference.

\paragraph{Frame classification without dictionary}
Following \citet{swayamdipta2017frame}, we also adopt the Lexical Unit (LU) dictionary in our model empirically. 
However, according to \citet{punyakanok08:importance}, sometimes the dictionary might be quite limited. 
Therefore, we offer one eaxmple in Table~\ref{tab:case:study} to illustrate the capability of our model for frames not in the dictionary. As shown, our model could predict the appropriate frame outside the dictionary as well and might additionally enrich the gold-standard annotations (i.e., the blue texts which do not appear in the Ground Truth).

\section{Conclusion}
In this paper, we proposed a novel graph-based model to address the end-to-end frame semantic parsing task. 
The full frame semantic parsing result of one sentence is organized as a graph,
and then we suggest an end-to-end neural model for the graph building.
Our model first encodes the input sentence for span representations with BERT, and then constructs the graph nodes and edges incrementally. To demonstrate the effectiveness of our method, we derived several pipeline methods and used them to conduct the experiments for comparisons. Experimental results showed that our graph-based model achieved significantly better performance than various pipeline methods. In addition, in order to compare our models with previous studies in the literature, we conducted experiments in the scenarios of the individual subtasks.
The results showed that our proposed models are highly competitive.

\section*{Acknowledgements}

We thank all reviewers for their hard work. This work is supported by grants from the National Key
Research and Development Program of China (No. 2018YFC0832101) and the National Natural
Science Foundation of China (No. 62176180).

\bibliography{custom}
\bibliographystyle{acl_natbib}
\end{CJK}
\end{document}